\documentclass[letterpaper,conference]{ieeeconf}

\IEEEoverridecommandlockouts                              

\usepackage{graphicx}
\usepackage{amssymb}  
\usepackage{hyperref,tcolorbox}
\usepackage{subcaption}
\usepackage{enumerate}
\usepackage{cite}
\usepackage{flushend}
\usepackage{amsmath,bm}
\usepackage[algo2e]{algorithm2e} 
\usepackage{comment}

\newcommand\norm[1]{\left\lVert#1\right\rVert}
\def\x#1{x_{#1}}
\def\y#1{y_{#1}}
\def\u#1{u_{#1}}

\def\d{d}

\def\rr#1{\mathbb{R}^{#1}}

\def\yhat#1{{\hat y}_{#1}}

\def\dh{\Delta h}
\def\ystar{y^*}
\def\du#1{\delta u_{#1}}
\def\ybar#1{{\bar y}_{#1}}
\usepackage{scrextend}

\newif\ifNotAnonymous
\NotAnonymoustrue

\title{\LARGE \bf
Learning and Adaptation in 
Wire Arc Additive Manufacturing Bead Geometry Control 
}

\ifNotAnonymous{
\author{Chen-Lung Lu$^\dagger$\thanks{$^\dagger$ Rensselaer Polytechnic Institute, Troy, NY. Emails: {\tt luc6@rpi.edu wenj@rpi.edu}}, John T.~Wen$^\dagger$
}
}\fi

\usepackage{titlesec}

\begin{document}
\maketitle
\thispagestyle{empty}
\pagestyle{empty}

\begin{abstract}

%
%
Robotics Wire Arc Additive Manufacturing (WAAM) is governed by complex and nonlinear process dynamics coupling thermal field to the build geometry. The process may be regarded as a multi-input/multi-output dynamical system with welding torch speed and wire feed rate as inputs and weld bead deposition height and width as outputs.  In this paper, we use the input/output data to learn a data-driven model and use it for weld planning and control.   We show that a simple recurrent neural network architecture and one-step-ahead predictive control can improve the process performance in terms of height and width consistency.  To account for the changing thermal conditions during the printing process, we update the learning model using prediction error from the previous layer.  This adaptation step further improves the prediction accuracy and controller performance. 
%
Experiments on a robotic WAAM testbed with integrated line-scanner feedback significant improvements in height and width consistency compared to constant input and static model baselines.  
The proposed learning and adaptation framework provides a practical pathway toward robust, data-driven regulation of additive manufacturing processes.
\end{abstract}
\noindent {\em Keywords:} 
Additive Manufacturing, WAAM, Data-Driven, Recurrent Neural Network, Predictive Control 

\section{INTRODUCTION}
\label{sec:intro}

Robotic Wire Arc Additive Manufacturing (WAAM) is a metal additive manufacturing process that builds near-net-shape components layer by layer using industrial robots for the torch motion.  WAAM offers high deposition rates and the ability to fabricate large-format structures, making it attractive for rapid prototyping, repair, and manufacturing of structural metal parts.
\begin{figure}[h!]
    \centering
    \includegraphics[width=0.9\linewidth]{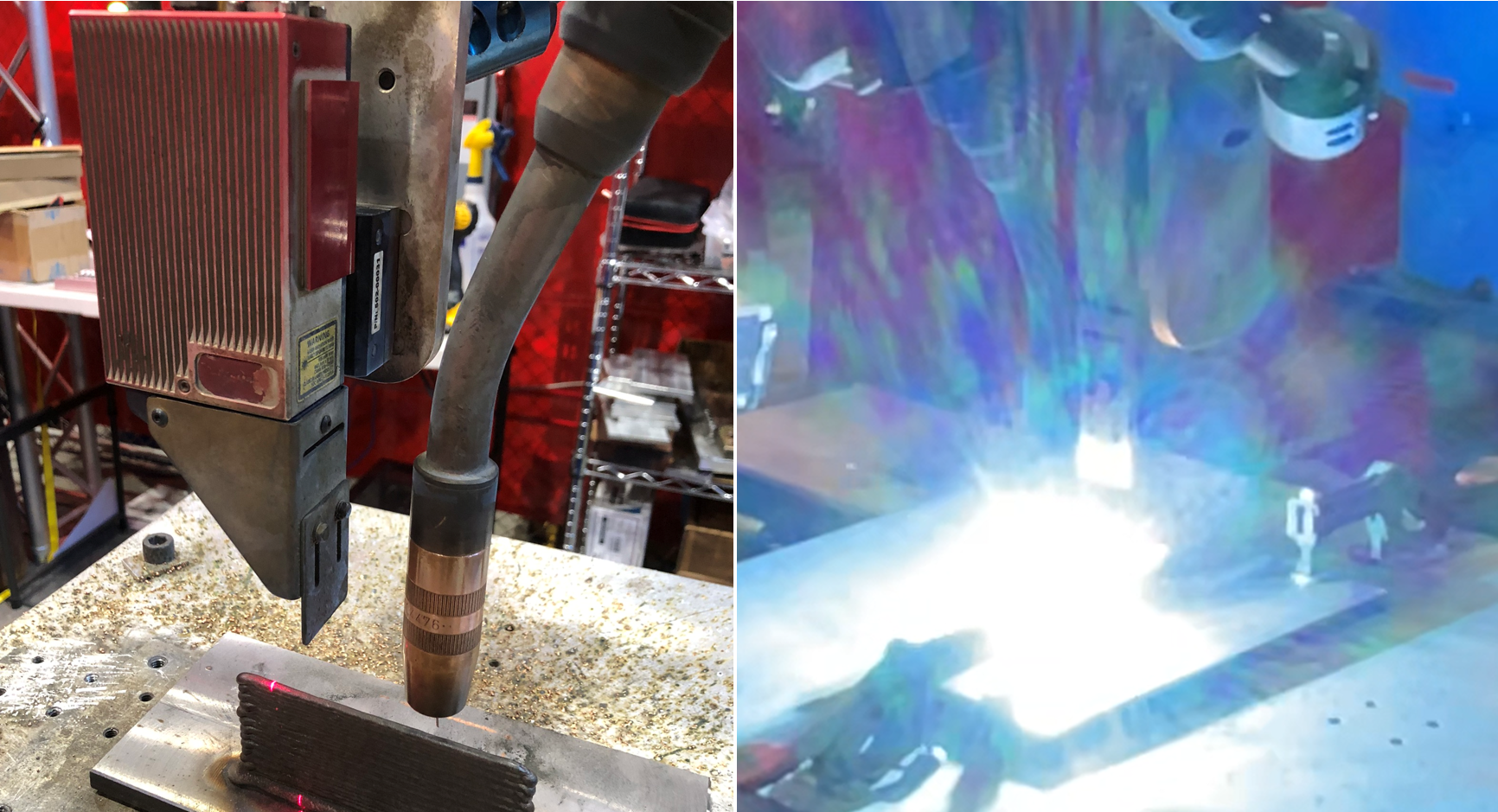}
    \caption{In-situ welding and sensing in a robotic WAAM process.}
    \label{fig:fujicam_weld}
\end{figure}
Despite these advantages, WAAM process suffers from defects such as geometric deformation, porosity, and cracking \cite{WU2018review}. Among these issues, geometric inaccuracy directly compromises the near-net-shape capability and increases post-processing requirements. 
Maintaining geometric stability is critical, as faulty bead geometry can accumulate through successive layers leading to unacceptable parts \cite{he2025open}. 

Bead geometry is governed by nonlinear thermomechanical dynamics involving multiple process variables, including torch travel speed, wire feed rate, welding current and voltage, and thermal conditions of the weld. 
Early analytical models describe deposition height and width using simplified static relationships between torch speed and wire feed rate and the deposition geometry \cite{Handa1997THERE}. However, such models neglect complex underlying dynamics, including heat transfer, melt pool formation, and material flow. 
%
More recently, data-driven approaches have been 
applied to relate process parameters to deposition geometry.
Deep neural networks (DNNs) have been used to predict layer height and width from process inputs, such as torch speed and wire feed rate \cite{so2024prediction}, welding current \cite{kaikui2023prediction}, welding voltage, nozzle to plate distance \cite{wacker2021geometry}, consecutive layers deposition and preheat temperature \cite{xiong2014bead}.  These methods are layer-by-layer approach, and does not provide in-layer prediction or control. 
As a result, the dynamics of the WAAM process are not  included. Furthermore, these studies often rely on relatively small datasets, limiting their ability to characterize transient process behavior. 

To capture the dynamic behavior of the WAAM process, input/output models such as autoregressive with exogenous inputs (ARX) have been used, in linear single-input/single-output (SISO)~\cite{xia2020model},
linear multi-input/multi-output (MIMO)~\cite{mu2022layer}, and nonlinear SISO~\cite{xiong2016closed} formulations. 
Using in-situ sensing such as cameras or line scanners, these approaches construct time-series datasets of torch speed, wire feed rate, and the resulting deposition height and width to identify system models.
For geometric control, these methods have employed PID control, model predictive control (MPC) based on the identified model, or adaptive single-neuron controllers. 
More recently, recurrent neural networks (RNNs) have been explored for modeling WAAM dynamics. In \cite{mendes2025data}, a Long Short-Term Memory (LSTM) network is used to model the dynamics from torch speed and wire feed rate to bead width. The trained model is subsequently approximated by a linear second-order system to guide the gain tuning of a PID controller. Other approaches consider learning multiple local linear models and tuning corresponding PID controllers~\cite{wang2022control}.
While these approaches demonstrate effectiveness, many ARX formulations rely on linear or locally linear input-output structures, or focus only on a single geometric output such as bead width or height. PID control has been used in welding and WAAM because of its simplicity and practical effectiveness. However, direct PID implementation for in-situ geometry control is ad hoc and requires careful tuning. 
Simultaneous height and width control is a coupled MIMO problem 
that requires tuning of a matrix gain, which also needs account to actuator bounds, input rate-change limit, or feasible deposition volume-per-distance (VPD). Recent studies have explored reinforcement learning (RL) for WAAM control. In~\cite{Xiong2019process, Dharmawan2020model}, Gaussian Processes (GPs) are used to model deposition behavior. The control strategy  selects the next input based on the learned GP model without considering a longer horizon. In~\cite{mattera2025optimal}, a reinforcement learning agent based on Deep Deterministic Policy Gradient (DDPG) is trained in a physics-based simulation; however, learning or validation on real WAAM datasets is not demonstrated. 

In this paper, we present an in-situ robotic WAAM control framework with torch speed and wire feed rate as input and bead height and width as output. We compare several common time series neural network architectures to model the deposition process within each layer.  These models are basically nonlinear state space systems containing networks of connected nodes and saturation functions. 
To train the parameters in these models, a dataset covering the feasible input space is collected.  The input/output process data are divided into the training set and evaluation set.  Network weights are adjusted using back propagation to best match the model output with the training data under the same input. The performance of various neural architectures using the evaluation dataset is similar.  Therefore, we choose the simplest architecture, simple RNN which contains one layer of sigmoid type of neurons, for the online implementation. 
During the WAAM geometry control, the input is updated to reduce the predicted output error in a one-step-ahead receding horizon control scheme using the learned model. This controller choice is motivated by the relatively smooth and stable nature of the WAAM geometry response and the need for computationally efficient online implementation. The experimental results show that repeated one-step corrections are effective in improving geometric consistency in the deposition task. In contrast to the previous work, our method learns a nonlinear recurrent input–output model and uses it directly in the control loop for simultaneous regulation of deposition height and width. 
To account for variations in thermal conditions and heat accumulation that are not fully represented in the offline training data, 
we fine-tune the trained model parameters based on the data from the immediately preceding layer.

The proposed scheme is evaluated on a robotic WAAM testbed with in-situ robot motion and welding control and deposition height and width measurement.  A rich dataset is acquired for a range of input conditions within the operating envelop to train and evaluate the models. 
Experimental results show that the RNN models achieve higher prediction accuracy than a static empirical model. Furthermore, adaptive fine-tuning using in-situ measurements from the preceding layer further improves the prediction accuracy, especially for deposition height.
In closed-loop deposition experiments, the simple RNN-based predictive control significantly outperforms constant input baselines and static input/output models in regulating bead height and width. The adaptive fine-tuning strategy further improves height consistency, although width regulation remains more sensitive to thermal fluctuation and coupled process effects.

The main contributions of this paper include:
\begin{enumerate}[1)]
    \setlength{\leftmargin}{0pt}  
    \setlength{\itemindent}{\labelwidth} 
    \item A data-driven framework that effectively captures the WAAM input–output dynamics using RNN models.
    \item The demonstration that a simple one-step-ahead predictive control strategy is sufficient for regulating WAAM deposition geometry using a trained simple RNN model.
    \item An adaptive fine-tuning strategy that uses prediction error from the preceding layer to improve the RNN model.
    \item A robotic WAAM testbed with in-situ sensing that enables large-scale dataset collection and experimental validation of closed-loop control performance.
\end{enumerate}
The paper is organized as follows. Section~\ref{sec:model_control} presents the data-driven modeling approaches, the one-step-ahead predictive control formulation and model adaptive fine-tuning. Section~\ref{sec:experiment} describes the robotic WAAM testbed and motion planning for welding and sensing and reports experimental results on prediction accuracy and closed-loop deposition quality.

\section{WAAM Modeling and Control}
\label{sec:model_control}

\subsection{WAAM Model}

As any additive process, WAAM is fundamentally a two-dimension system consisting of the in-layer dynamics and layer-by-layer evolution.  The in-layer dynamics maps the input torch speed, $v_T$ and wire feed rate $v_W$ to the output deposition height $\dh$ and width $w$ through nonlinear dynamics as represented in a state space model below (for the $i$th layer):
\begin{equation}
    \x{k+1}^{(i)} = f(\x k^{(i)},\u k^{(i)}), \quad \y k ^{(i)} = f_o( \x k^{(i)}) 
    \label{eq:nonlindyn}
\end{equation}
where $\u k^{(i)} = \begin{bmatrix} v_{T_k}^{(i)}\\ v_{W_k}^{(i)} \end{bmatrix}$, $\y k^{(i)}= \begin{bmatrix} \dh_k^{(i)}\\ w_k^{(i)} \end{bmatrix}$, $\x k\in\rr n$ is the state with dimension $n$, and $f$ and $f_o$ for the state evolution map and output map, respectively. 
The layer height accumulates additively at the path location $s$:
\begin{equation}
    h^{(i)}(s) = h^{(i-1)}(s) + \dh^{(i)}(s)
    \label{eq:layerheight}
\end{equation}
where $s$ is the path length variable and $h^{(i)}(s)$ is the height of the build of the $i$th layer at the $s$. The temporal $\dh_k$ and spatial path $\dh(s)$ are approximately related via
\begin{equation}
    s_{k+1}=s_k + v_{T_k} \, t_s, \quad s_0=0,
\end{equation}
where $t_s$ is the sampling period.  This relationship is not exact due to the discretization of $v_{T}$.
The WAAM process dynamics is primarily governed by heat transfer.  The heat input melts the wire to form a molten pool of liquefied metal.  The shape of the weld bead (width and height) is determined by the thermal state of the molten pool.  Therefore, the state vector is likely related to the thermal state of the build at the $i$th layer and the state dynamics is governed by the thermomechanical process. This process is inherently stable as heat will eventually dissipate into the environment.  A bounded input will result in bounded output, hence the physical system is bounded-input/bounded-output (BIBO) stable.  As the layers accumulate, so does heat in the lower layers.  Hence, we expect cooling rate is lower at the higher layers.  There is also the additional complication of arc on/off effect at the start and end of each layer which tend to deposit more at the start and less at the end.  The $\dh$ measurement may contain a spatial offset (as in our testbed).  This would result in a delayed measurement that will need to be compensated.  Together, these effects would compromise model learning and control at the edge of each layer.

\subsection{Control Objective}
The first step in additive manufacturing is to {\em slice} the target parts into layers.  This will generate the target height ${h^{(i)}_k}^*$.  We will focus on the {\em thin wall} case (single-bead build), so the target width is a constant $w^*$.
%
For the $i$th layer, the control objective is to choose $\u k^{(i)}$ to steer $\dh_k^{(i)}$ to ${h^{(i)}_k}^*-h_k^{(i-1)}$ and $w_k^{(i)}$ to $w^*$.  Current practice in WAAM mostly chooses $\u k^{(i)}$ as a constant input for the layer.  The value may change by layer as the thermal condition may be different.  There also may be scanning between layers to obtain $h_k^{(i-1)}$ to adjust the deposition setpoint for the $i$th layer.  This scan-n-print approach has been used in \cite{lu2025multi} using a static model from $v_T^{(i)}$ to $\dh^{(i)}$.  As the system is BIBO stable, our focus is on the performance, i.e., minimizing the difference between $h_k^{(i)}$ and ${h^{(i)}_k}^*$. 

\subsection{Approximate Neural Models}
\label{sec:model_training}

We will train a neural model to approximate \eqref{eq:nonlindyn} using experimental input/output data.  The approximate model will then be used to generate the control input in each layer based on the real-time measurements of the deposit. 
For the in-layer control, we will drop the superscript $\,^{(i)}$ for clarity. We consider several common recurrent neural networks, including simple RNN, Long-Short Term Memory (LSTM), Gated Recurrent Unit (GRU) \cite{simpleRNN,lstm,gru}.  They all fit into the same structure as \eqref{eq:nonlindyn} with $f_o$ as an affine function:
\begin{equation}
    f_o(\x k) = C \x k + \d .
\end{equation}
For comparison, we also consider the neural network implementation of the nonlinear ARX model (NN-NARX) which relates $y_k$ to past inputs and outputs. 
The full expression of all these models are included in Appendix~\ref{app:RNN}. 
The RNN, LSTM, and GRU models are trained using time series input/output data and optimized via backpropagation through time.  The NN-NARX model is trained using the standard backpropagation. The models are trained using mean square error as the loss function and updated with Adam optimizer \cite{Kingma2014AdamAM}. As the physical process is BIBO stable, the trained models need to retain the same stability property.

\subsection{Static Baseline model}


%
It has been noted that the deposition height is related to the torch speed via a power law \cite{Handa1997THERE}.  
The constants in the relationship may be estimated through a least square fit by applying the logarithmic function to both sides. 
%
%
This relationship has been used in \cite{lu2025multi} for layer-by-layer height correction. 
We extend this log-log model to relate $(v_T,v_W)$ to $(\dh,w)$, with the constants estimated through a least-square fit: 
\begin{equation}
    \begin{aligned}
        \ln(\Delta h) &= \alpha_0 \ln(v_T) + \alpha_1 \ln(v_W) + \alpha_2 \\
        \ln(w) &= \beta_0 \ln{v_T} + \beta_1 \ln(v_W) + \beta_2.
    \end{aligned}
    \label{eq:loglog}
\end{equation}



\subsection{One-Step-Ahead Predictive Control}
\label{subsec:one_step_ls_formulation}

Our control strategy is to choose $\u k$ to move the predicted one-step-ahead output $\yhat{k+1}$ closer to the target output $\ystar_{k+1}$ than $\y k$.  
For the $i$th layer, 
\begin{equation}
    y^{(i)^*}_{k} =\begin{bmatrix}
        h^{(i)^*}(s_k^{(i)}) - h^{(i-1)}(s_k^{(i)}) \\ w^*
    \end{bmatrix}
\end{equation}
where $s_k^{(i)}$ is the $k$th spatial point along the $i$th layer path, $h^{(i)^*}(s_k^{(i)})$ is the target height 
$s_k^{(i)}$, and $h^{(i-1)}(s_k^{(i)}$ is based on the interpolation of recorded $h_k^{(i-1)}$ and $s_k^{(i-1)}$ to match with the location $s_k^{(i)}$.  
In our experiments, we use back-and-forth printing, so care must be taken to ensure the path indices are aligned properly (i.e., data from the previous layer needs to be flipped).
Using \eqref{eq:nonlindyn}, $\y{k+1}$ may be expressed as
\begin{equation}
    \y{k+1} = C f(\x k,\u k) + \d.
\end{equation}
At time $k$, $\x k$ is known. Let $\u k$ be a small increment from the known input at the previous time step $\u{k-1}$:
\begin{equation}
    \u{k} = \u{k-1} + \du k.
\end{equation}
Expand $\hat f$ about $\u{k-1}$, we have 
\begin{equation}
\begin{split}
    \y{k+1} \approx
    \underbrace{C f(\x k,\u{k-1},\y k-\y k) + \d}_{\ybar{k+1}} \\
    + \underbrace{C \nabla_u f(\x k,\u{k-1})}_{J_k} \du k
\end{split}
\end{equation}
where $\ybar{k+1}$ and $J_k$ are both known.  

At each time step $k$, we seek a control input $\du k$ that minimizes the one-step-ahead predicted tracking error 
\begin{equation}
    \min_{\du k} \norm{W(J_k \du k + \bar y_{k+1} - \ystar_{k+1})}^2 + \lambda ||\delta u_t||^2
    \label{eq:control}
\end{equation}
subject to the input and rate constraints
$$
    \du {\min} \le \du k \le \du {\max}, \quad
    \u{\min} \le \u k + \du k \le \u {\max},
$$
where $W=\mathrm{diag}(w_{\Delta h},\, w_w)$ is a weighting matrix, $\Lambda = \mathrm{diag}(\lambda_{\Delta h},\lambda_w)$, $\lambda_{\Delta h},\lambda_w \geq0$ is a regularization term, $(\du{})_{\min/\max}$ are the control inputs rate limits, and $(\u{})_{\min/\max}$ enforce physical bounds on torch speed and wire feed rate. 

To avoid overly aggressive control which may lead to thermal issues, we modify the input update with a step size:
\begin{equation}
    \u k = \u{k-1} + \alpha \du k
    \label{eq:control_update}    
\end{equation}
where $\alpha \in (0,1]$ is a step size.  To meet online computation constraints, we adopt a real-time iteration strategy and perform a single optimization update per time step, warm-started from the previous control input.

\subsection{Adaptive Model Fine-Tuning}
\label{sec:adaptive_update}

WAAM process dynamics may vary due to changes in thermal conditions, heat accumulation, and layer-dependent process behavior that are not fully captured by the offline training dataset. To account for these variations, the in-situ sensing system can be used to collect newly deposited layer data and adaptively fine-tune the learned model during welding. The fine-tuning procedure follows the same training strategy described in Section~\ref{sec:model_training}. However, instead of using the full offline training dataset, the model is updated using data collected from the previously deposited layers in the current build. To prevent the adapted model from drifting excessively from the original model, a regularization term is added to the mean-squared prediction loss:
\begin{equation}
    \mathcal{L}_i =
    \left\|\hat{y}_{i} - y_{i}\right\|^2
    +
    \lambda
    \left\|\theta - \theta_{i}\right\|^2 ,
\end{equation}
where $\hat{y}_{i}$ is the model prediction for layer $i$, $y_i$ is the corresponding measured output, $\theta$ denotes the model parameters being optimized during fine-tuning, $\theta_i$ denotes the model parameters before the adaptive update, and $\lambda$ is the regularization weight. This regularized fine-tuning strategy allows the model to adapt to newly observed process behavior while limiting excessive parameter changes, thereby improving robustness during adaptive control.

\section{Implementation and Evaluation}
\label{sec:experiment}
\subsection{Robotic WAAM Testbed}
\label{sec:testbed}

\begin{figure}
    \centering
    \includegraphics[width=0.8\linewidth]{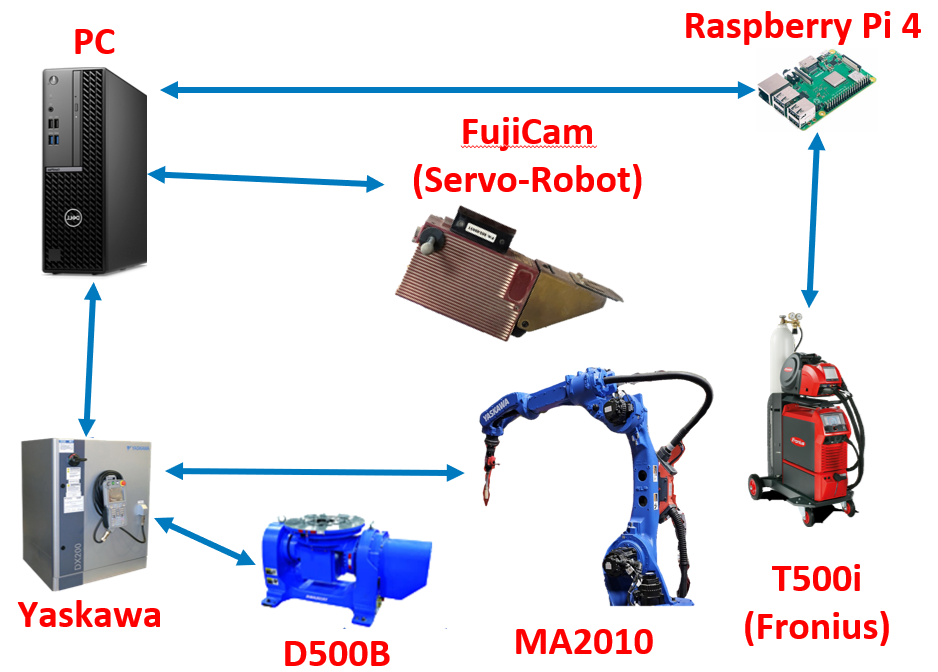}
    \caption{Robotic WAAM testbed showing the robot arm, co-located sensors, welding equipment, and computing hardware.}
    \label{fig:waam_hardware}
\end{figure}

Our WAAM system comprises a Yaskawa MA2010 industrial welding robot with a Servo Robot FujiCam line scanner, a two degree-of-freedom (dof) D500B trunnion table, and a Fronius TPS500i welding power source. All devices are connected to a centralized PC for coordination and data logging.  
The system sends robot joint setpoint commands to the robot at 125~Hz and receives joint angle measurements at 250~Hz using the MotoPlus streaming interface. The line scanner acquires line profiles at 250~Hz, and the wire feed rate is commanded at 10~Hz. Control commands and sensor data are exchanged through custom RobotRaconteur drivers, enabling synchronized streaming and coordinated operation across the system. 
The overall system architecture has been described in \cite{he2025open} and is shown in Figure~\ref{fig:waam_hardware}.


\subsection{WAAM Model Training Dataset}

\begin{figure}[ht]
    \centering
    \includegraphics[width=0.35\textwidth]{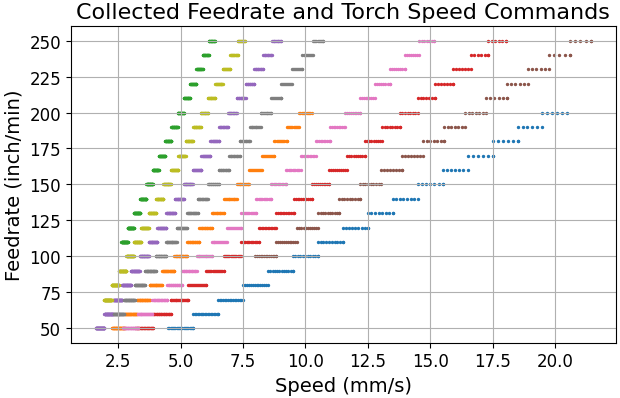}
    \caption{Coverage of collected torch speed and wire feed rate combinations in the WAAM dataset.}
    \label{fig:data_speed_feedrate}
\end{figure}

All experiments in this study are conducted using ER316L stainless steel, a commonly used material in WAAM processes. 
A dataset mapping WAAM command inputs to geometric deposition outputs is collected using a 110~mm long thin-wall geometry. In total, data are collected across 21 distinct wire feed rates and 2765 torch speed settings, spanning a wide range of operating conditions relevant to online control.
Figure~\ref{fig:data_speed_feedrate} shows the coverage of the collected dataset in the torch speed vs. wire feed rate space. Each color corresponds to data collected from the same thin-wall build, which shares a similar volume-per-distance (VPD $= \frac{v_W}{v_T}$) condition. For each experiment, command inputs and geometric measurements are synchronized and recorded at the achievable control sampling rate 10 Hz, resulting in time-series trajectories suitable for training and evaluating input--output dynamical models. The dataset is randomly split into training and validation sets using an 80/20 ratio.

\subsection{Model Training and Prediction}







We train the four common model structures, RNN, LSTM, GRU, and NN-NARX. 
The models were implemented and trained in Python using PyTorch~\cite{pyTorch} with CUDA backend acceleration.

To evaluate the sensitivity of model performance to model capacity, we conduct an ablation study by training models with hidden sizes or history lengths of $\{3, 8, 16, 64\}$. 
%
%
All models are trained using mean squared error (MSE) as the loss function and optimized using the Adam optimizer. Each model is trained for 5000 epochs. Table~\ref{tab:forward_time} shows the average single-step forward inference time for all model structures and system orders. All models exhibit comparable per-step latency, indicating that for these dimensions the measured computation time is dominated by PyTorch execution overhead and GPU kernel launch latency rather than arithmetic cost.

\begin{table}[ht]
\begin{center}
\begin{tabular}{ c c c c c } 
\hline
\hline
 & RNN & GRU & LSTM & NARX  \\
\hline
3  & 0.1396 & 0.1714 & 0.2001 & 0.1213 \\
8  & 0.1382 & 0.1556 & 0.1699 & 0.1258 \\
16 & 0.1408 & 0.1592 & 0.1608 & 0.1198 \\
64 & 0.1804 & 0.1551 & 0.1433 & 0.1246 \\
\hline
\hline
\end{tabular}
\caption{Average single-step forward inference time for different model structures and hidden sizes or history lengths. Unit: ms}
\label{tab:forward_time}
\end{center}
\end{table}

\begin{figure}[ht]
    \centering
    \includegraphics[width=0.30\textwidth]{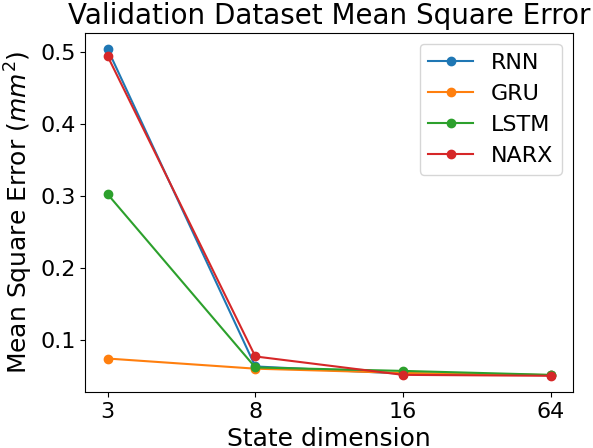}
    \caption{Validation mean squared error for different model structures and hidden sizes or history lengths.}
    \label{fig:validation_loss}
\end{figure}

The validation loss for different model structures and capacities are summarized in 
Figure~\ref{fig:validation_loss}. 
Increasing the hidden state dimension or history length improves prediction accuracy, although the improvement saturates beyond a size of 16. Overall, the prediction performance does not vary significantly across different model architectures.  

\begin{figure}[htbp]
     \centering
     \begin{subfigure}[b]{0.235\textwidth}
         \centering
         \includegraphics[width=\textwidth]{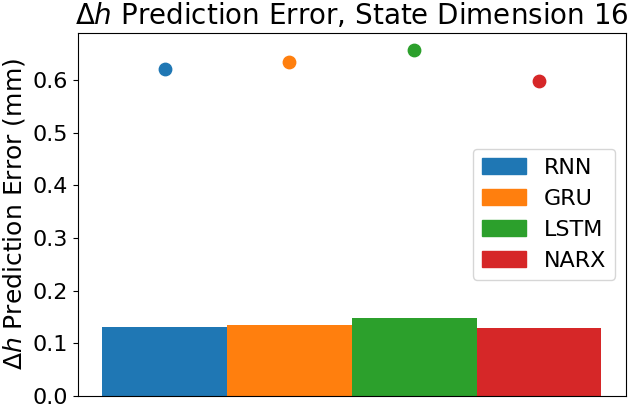}
         \caption{$\Delta h$ prediction error.}
         \label{fig:height_train_error}
     \end{subfigure}
     \begin{subfigure}[b]{0.235\textwidth}
         \centering
         \includegraphics[width=\textwidth]{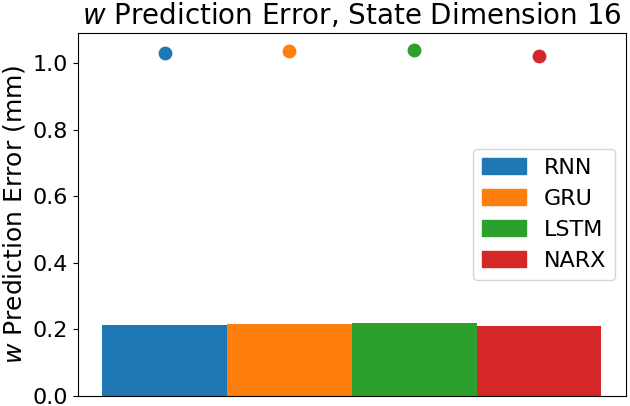}
         \caption{$w$ prediction error.}
         \label{fig:width_train_error}
     \end{subfigure}
    \caption{Deposition height ($\Delta h$) and width ($w$) absolute prediction errors for different model structures with state dimension 16. Mean (the bars), 95th-percentile (the dots) are reported.}
    \label{fig:height_width_error}
\end{figure}

We also compare the mean absolute error, 95th-percentile error for deposition height and width in Figure~\ref{fig:height_width_error}, using the state dimension  of 16. 
Large deposition height prediction errors primarily occur at the beginning and end of the welding beads, corresponding to transient process states. Large deposition width errors are observed at bead start and end regions as well as under high wire feed rates and torch speeds. As the performance is similar for all four architectures, we just use the simple RNN architecture for online implementation. 


\begin{figure*}[hbtp]
    \centering
    \begin{subfigure}[b]{0.24\textwidth}
        \centering
        \includegraphics[width=\textwidth]{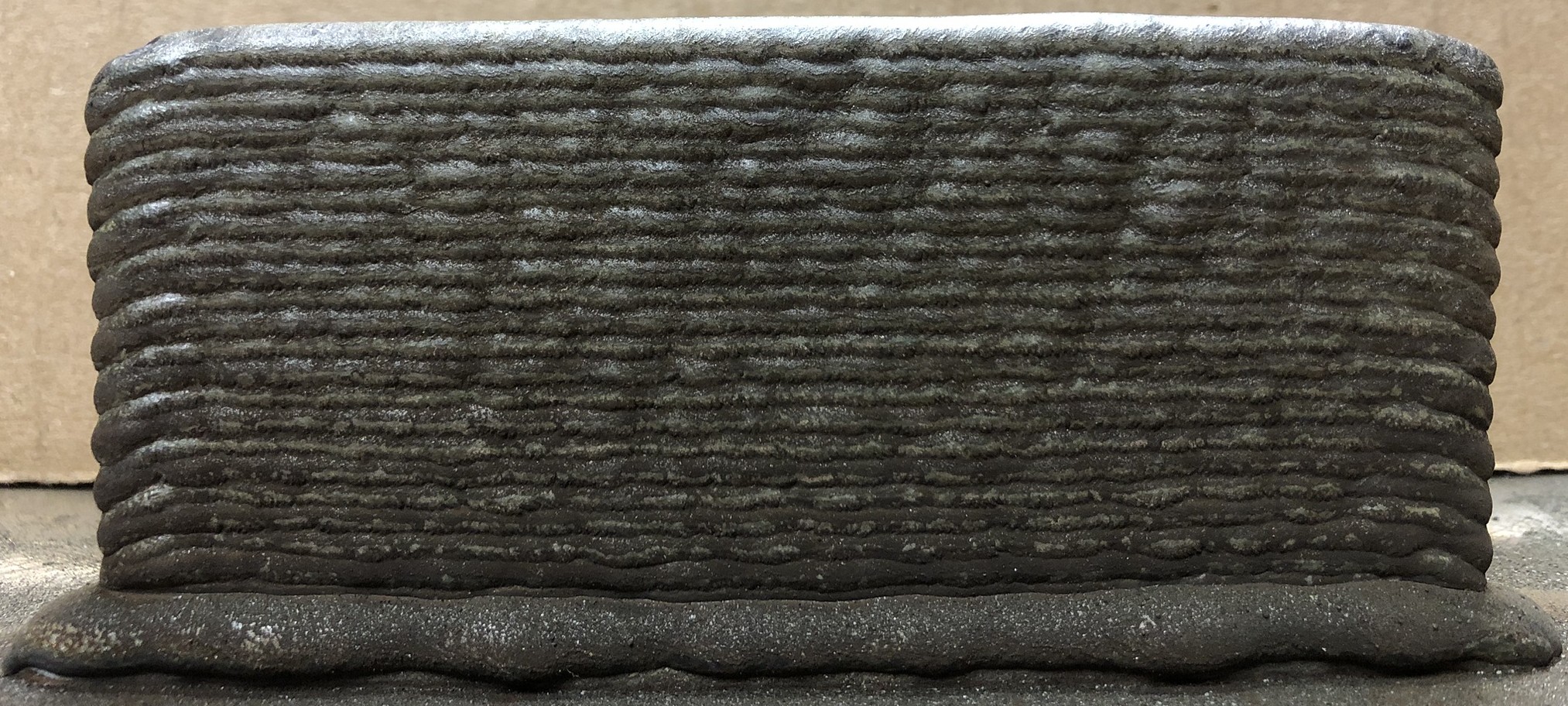}
        \caption{Baseline}
        \label{fig:weld_baseline}
    \end{subfigure}
    \begin{subfigure}[b]{0.24\textwidth}
        \centering
        \includegraphics[width=\textwidth]{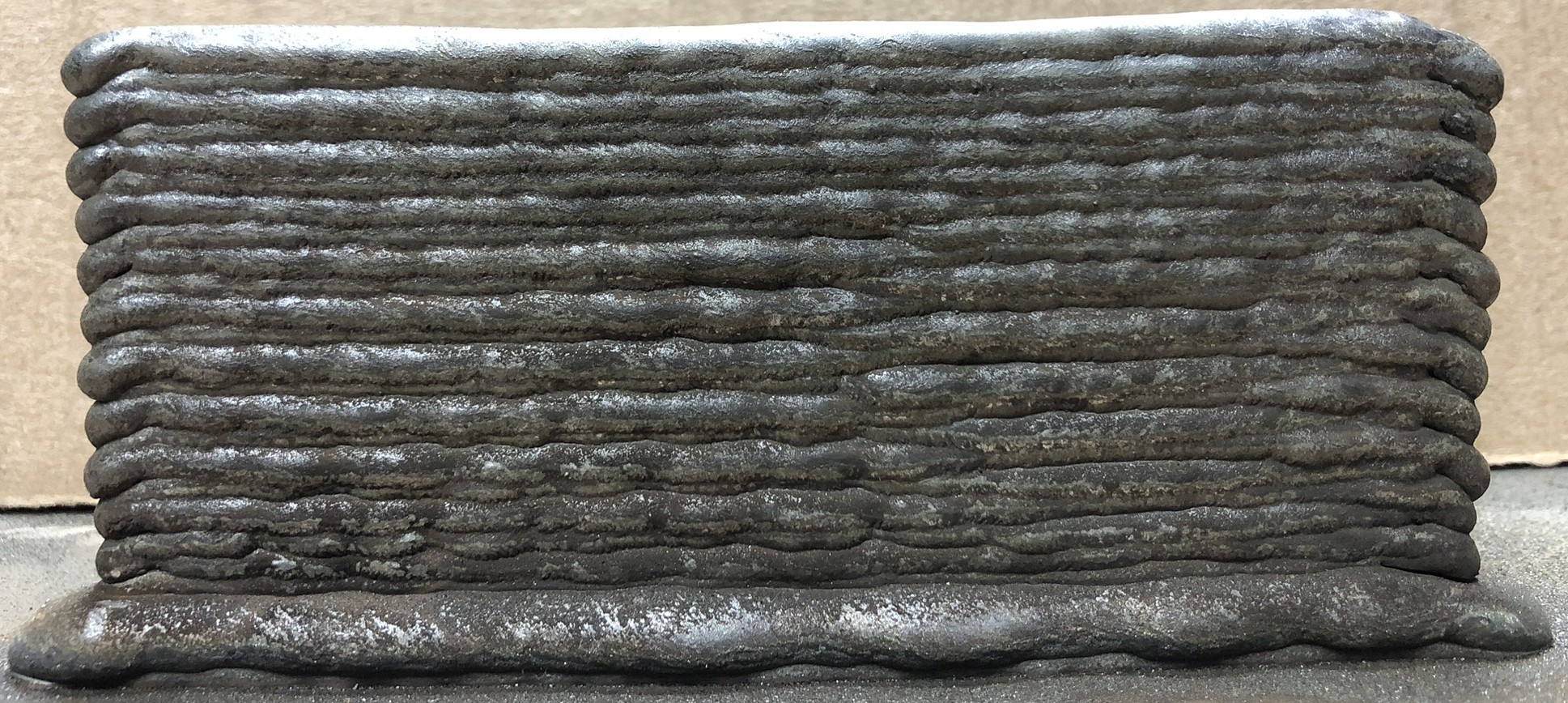}
        \caption{Log-log}
        \label{fig:weld_loglog}
    \end{subfigure}
    \begin{subfigure}[b]{0.24\textwidth}
        \centering
        \includegraphics[width=\textwidth]{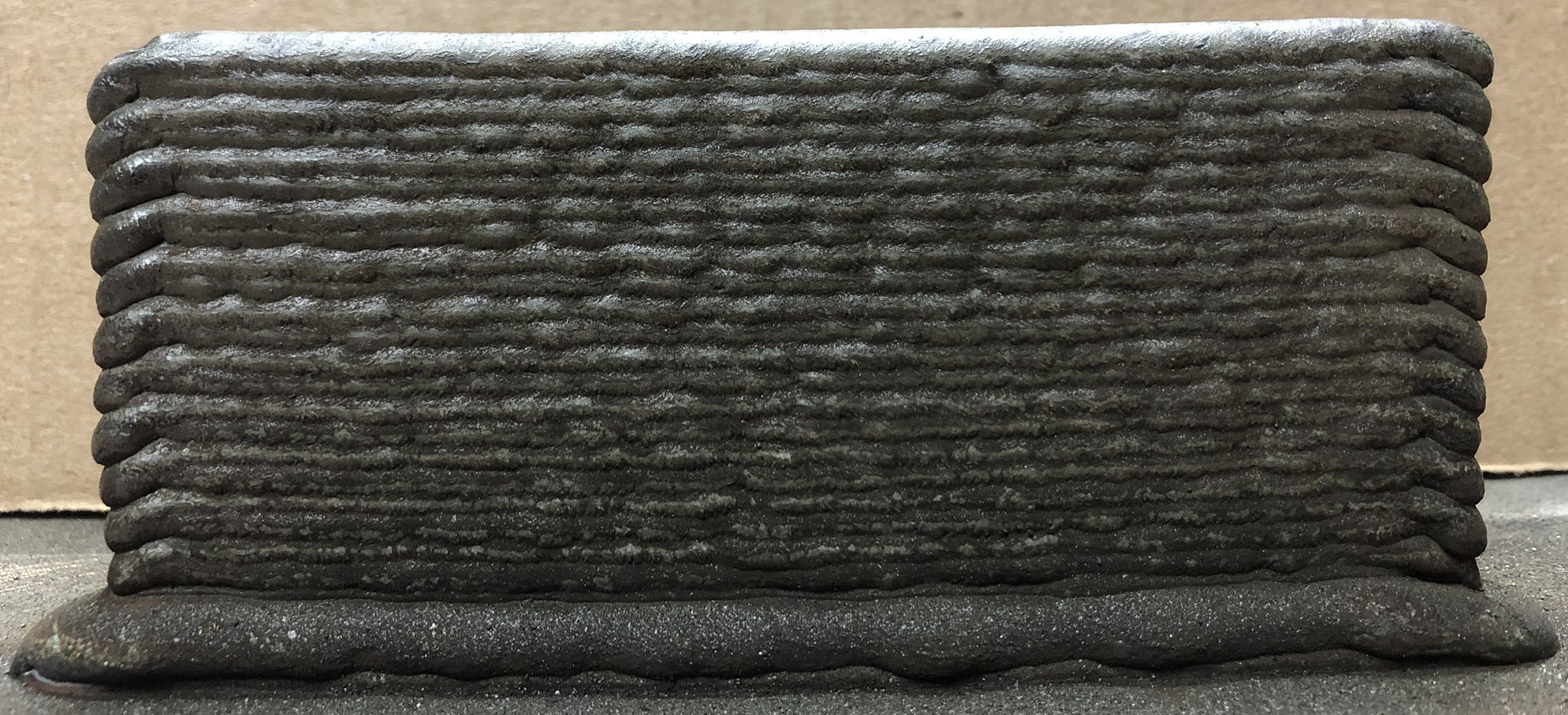}
        \caption{RNN}
        \label{fig:RNN_open}
    \end{subfigure}
    \begin{subfigure}[b]{0.24\textwidth}
        \centering
        \includegraphics[width=\textwidth]{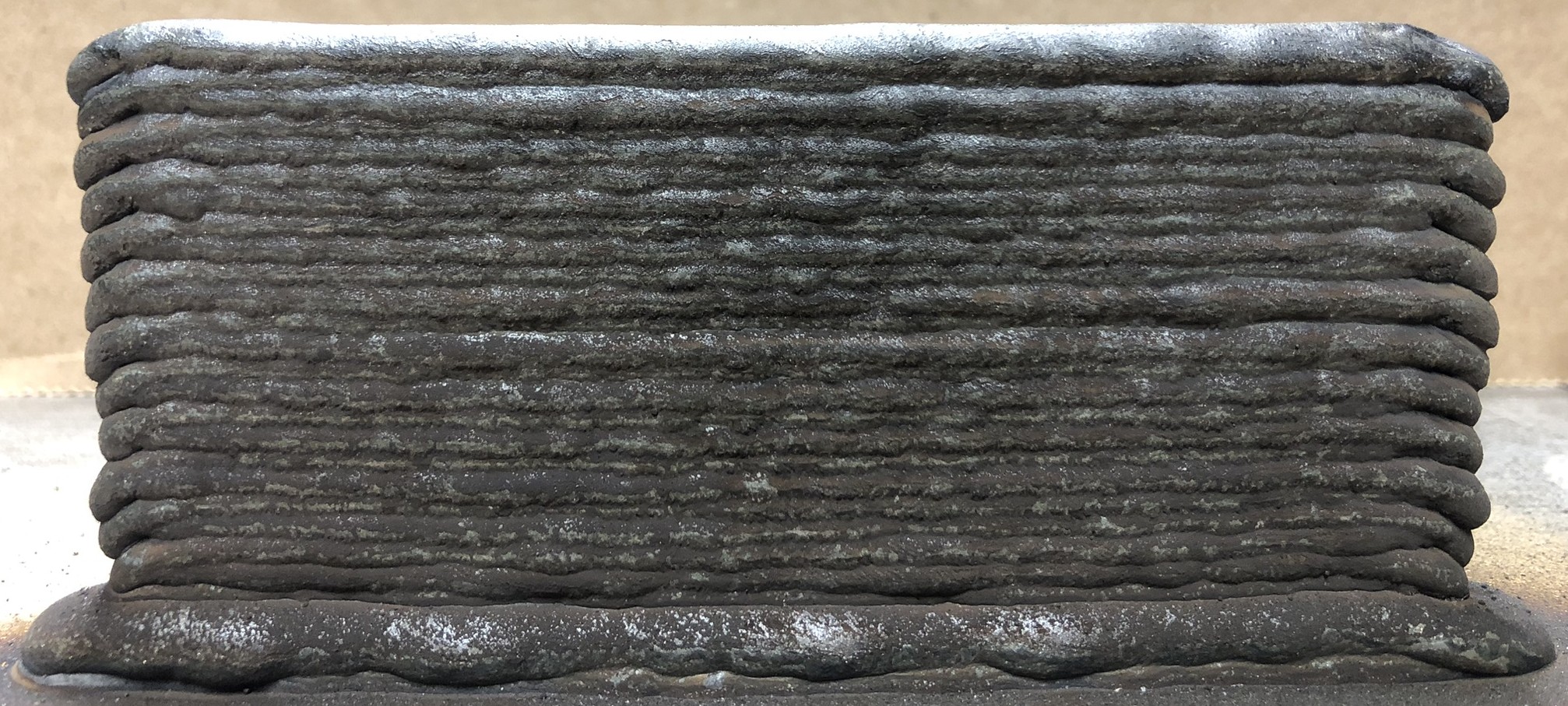}
        \caption{RNN-Adaptive}
        \label{fig:RNN_adaptive}
    \end{subfigure}
    \caption{Rectangular weld pieces produced under the four control implementations.}
    \label{fig:weld_peices}
\end{figure*}

\subsection{WAAM Height and Width Control}


\subsubsection{Controller Implementation and Parameter Settings}

We evaluate closed-loop geometric control using the trained models and the one-step-ahead least square error formulation introduced in Section~\ref{subsec:one_step_ls_formulation}. The target geometry is a rectangular wall of length $110$~mm, width $5.2$~mm, and height $50$~mm. The objective is to produce a geometry as close to an ideal rectangular wall as possible. Height and width standard deviation are used as indicators of deposition smoothness and geometric consistency.  Note that the torch motion is back and forth meaning that the printing reverses direction after each layer.  The torch is turned on at the start of the layer and turned off at the end of the layer to allow all acquired data to be transmitted from memory to disk storage.  

We compare four control implementations:
\begin{enumerate}
    \item \textbf{Baseline}: Constant torch speed and wire feed rate.
    \item \textbf{Log–Log Model}: Using the inversion of the static log–log model \eqref{eq:loglog} to compute the control input.
    \item \textbf{RNN}: Using the trained RNN model \eqref{eq:nonlindyn}   with the one-step-ahead predictive control \eqref{eq:control}-\eqref{eq:control_update}.
    \item \textbf{RNN-Adaptive}: Using the trained RNN model with adaptive fine-tuning based on previous-layer data, as described in Section~\ref{sec:adaptive_update}, together with the one-step-ahead predictive control \eqref{eq:control}--\eqref{eq:control_update}.
\end{enumerate}
We choose the nominal control inputs to be at the center of the data collection range, namely torch speed $v_T=7.5$ mm/s and wire feed rate $v_W=63.5$ mm/s. The baseline case then uses these values as constant torch speed and wire feed rate. The target deposition height and width are defined as the steady-state values obtained by applying these inputs to the trained RNN model.
\begin{equation}
    y^* = \!\!\!\begin{bmatrix}
        \Delta h^*  \\ w^* 
        \end{bmatrix}
       \! = \!\begin{bmatrix}
        1.8 \\
        5.2 
    \end{bmatrix}\!\!\mbox{mm},    
    \,\,u^* = \!\!\!\begin{bmatrix}
        v_T^*\\ v_W^* \end{bmatrix}
        \!=\!
        \begin{bmatrix}
        7.5 \\
        63.5
    \end{bmatrix}\mbox{mm/s}.
\end{equation}
%
%
%
The one-step-ahead predictive control uses the trained simple RNN model with state dimension 16 based on the ablation study. The regularization weights are chosen to be $0.01$ and $0.1$ for $v_T$ and $v_W$ increments to penalize changes in $v_W$ more heavily, as the wire feed rate has larger quantization steps.  
%
The maximum change per control step is $2$~mm/s for $v_T$ and 
4.23~mm/s
for $v_W$. These values are chosen to avoid rapid change per control step which could lead to undesirable builds. They are tuned in simulation using the simple RNN model.
Bounds for $v_T$ are set 2–15~mm/s and 
21.2-105.8~mm/s for $v_W$.
These bounds are chosen for safety and  to avoid inconsistent operations. 
If the volume per distance (VPD), $v_W/v_T$, is too high, there will be excessive melt.  If VPD si too low, the deposit will not be sufficient to achieve the desired geometry. 
The step size parameter $\alpha$ is set to 0.3. This is also tuned in simulation to avoid large overshoots in the output response. Prediction and control updates are performed at 10~Hz, corresponding to the update rate of the wire feed rate controller. To maintain synchronization between control variables, torch speed commands are updated at the same rate. In addition, to ensure consistent thermal conditions across all layers and runs, we set an inter-layer wait time of $45$ second. The parameters used in the control experiments are summarized in Table~\ref{tab:mpc_params}.


\begin{table}[ht]
\centering
\begin{tabular}{l c}
\hline
\hline
Parameter & Value \\
\hline
Target layer height $\Delta h^*$ & 1.8 mm \\
Target bead width $w^*$ & 5.2 mm \\
Torch speed limits & 2–15 mm/s \\
Wire feed rate limits & 21.2–105.8 mm/s \\
Max $\Delta$ $v_T$ per step & 2 mm/s \\
Max $\Delta$ $v_W$ per step & 4.23 mm/s \\
Regularization weight (torch speed) $\lambda_{\Delta h}$ & 0.01 \\
Regularization weight (wire feed rate) $\lambda_{w}$ & 0.1 \\
Step size $\alpha$ & 0.3 \\
RNN hidden size & 16 \\
Control / prediction update rate & 10 Hz \\
Interlayer wait time & 45 sec \\
\hline
\hline
\end{tabular}
\caption{Parameters used in one-step-ahead predictive control experiments.}
\label{tab:mpc_params}
\end{table}

\begin{figure}[htbp]
     \centering
     \begin{subfigure}[b]{0.235\textwidth}
         \centering
         \includegraphics[width=\textwidth]{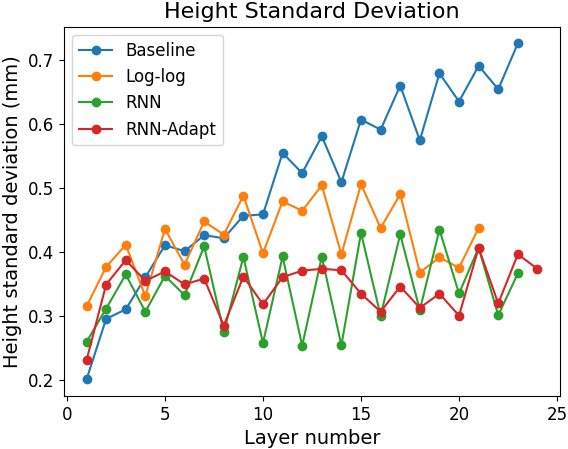}
         \caption{Height SD.}
         \label{fig:hstd_edge}
     \end{subfigure}
     \begin{subfigure}[b]{0.235\textwidth}
         \centering
         \includegraphics[width=\textwidth]{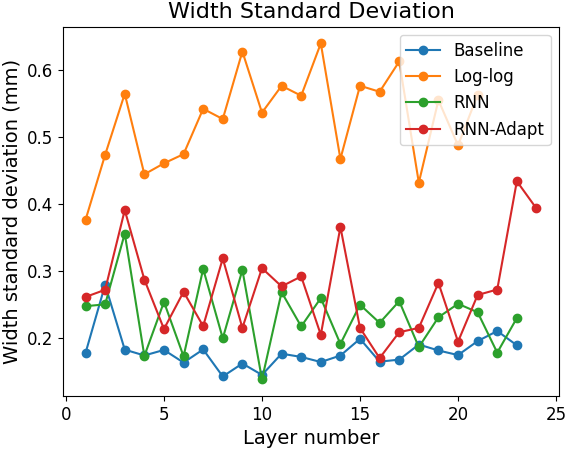}
         \caption{Width SD.}
         \label{fig:wstd_edge}
     \end{subfigure}
     \begin{subfigure}[b]{0.235\textwidth}
         \centering
         \includegraphics[width=\textwidth]{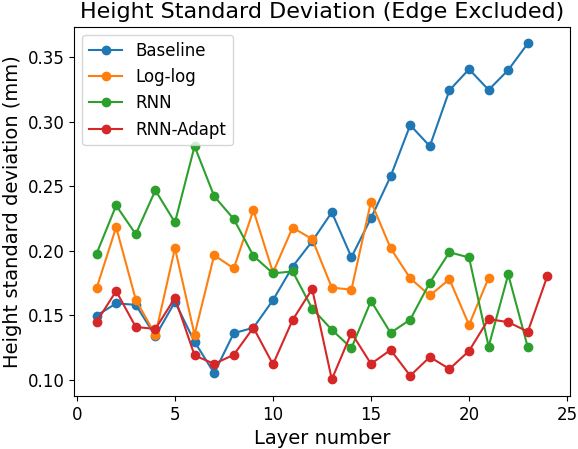}
         \caption{Height SD. Edge Excluded.}
         \label{fig:hstd_edge_excluded}
     \end{subfigure}
     \begin{subfigure}[b]{0.235\textwidth}
         \centering
         \includegraphics[width=\textwidth]{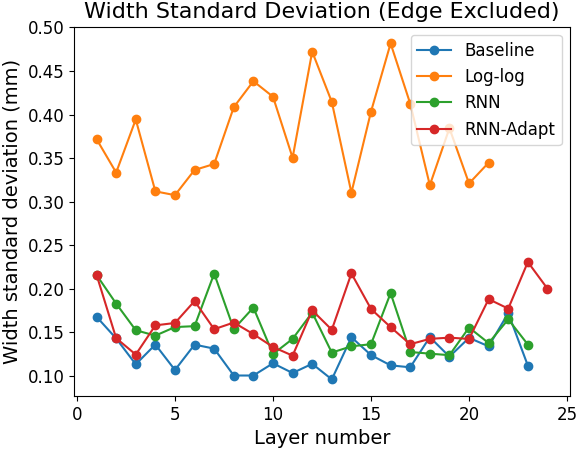}
         \caption{Width SD. Edge Excluded.}
         \label{fig:wstd_edge_exclided}
     \end{subfigure}
    \caption{Height and width standard deviation across layers for the four control implementations, with and without edge regions.}
    \label{fig:height_width_control_std}
\end{figure}


\subsubsection{Control Performance}

Figure~\ref{fig:weld_peices} shows the resulting weld pieces under the four control implementations. From visual inspection, the baseline controller exhibits a noticeably curved height profile compared to the other methods. This behavior is attributed to heat accumulation near the edges of the wall, results in lower but wider built. For the log–log controller, the overall height control is relatively consistent; however, the bead width control is less satisfactory. 
Both RNN and RNN-Adaptive controllers produce visually uniform height and width profiles.

Figure~\ref{fig:height_width_control_std} presents the height and width standard deviation (SD) across layers for all control implementations. The baseline height SD increases rapidly, reflecting the curvature observed in the weld profile. 
In contrast, the log–log, RNN model controllers maintain bounded height SD, with RNN-Adaptive showing the most consistent performance. For width smoothness, the log–log controller shows higher SD, indicating limited width control despite reasonable height regulation. Both RNN model controller achieves low and stable width SD while simultaneously maintaining height control. The RNN-Adaptive has a slightly better height SD, but slightly higher width SD.

We also analyze WAAM performance with edge regions excluded from the SD computation, as these correspond to transient welding phases. The edges are defined as $5$~mm within the welding arc on and off points.
Even with edge regions excluded, the baseline controller continues to show increasing height non-uniformity as the wall builds. The RNN-based controller maintains low and stable height SD. Adaptive fine-tuning further reduces the height SD.
For width profiles,
the baseline method appears even smoother than when edges are included, since static inputs typically cause excess deposition at the start and reduced deposition at the end of the weld. Removing these transient regions reveals a more stable central width profile. This behavior also reflects the inherent geometric deposition stability of ER316L stainless steel under steady welding conditions. However, the log–log controller still exhibits relatively large width variation, whereas the RNN-based controllers maintain both low and stable width SD.
\begin{table}[ht]
\begin{center}
\begin{tabular}{ c c c c c } 
\hline
\hline
 & Height & Width & Height (EX) & Width (EX)  \\
\hline
Baseline & 0.51 & 0.18 & 0.22 & 0.13 \\
Log-log & 0.42 & 0.53 & 0.18 & 0.38 \\
RNN & 0.34 & 0.23 & 0.18 & 0.16 \\
RNN-Adaptive & 0.34 & 0.27 & 0.13 & 0.16 \\
\hline
\hline
\end{tabular}
\caption{Average height and width standard deviation across all layers. EX denotes edge-excluded results. Unit: mm}
\label{tab:hw_std}
\end{center}
\end{table}
Table~\ref{tab:hw_std} summarizes the average SD values. Both RNN-based controllers outperform the log--log approach. Although the baseline case produces the smoothest width profile, it eventually leads to a significantly tilted height profile.

\subsection{Model and Controller Characteristics}


When the torch first turns on, there tends to be a large deposit from the initial melting of the wire. 
When the torch turns off at the end of a layer, there is a reduced melt pool and the deposit drops sharply.
The back and forth printing tends to balance out these two effects. This explains the sawtooth shape of the height and width variations in each layer as seen in Figure~\ref{fig:height_width_control_std}.

\begin{figure}[h!]
\centering
    \begin{subfigure}[b]{0.235\textwidth}
    \centering
    \includegraphics[width=\textwidth]{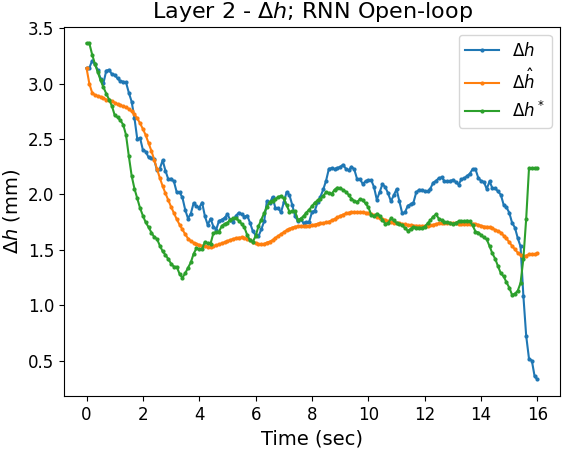}
    \caption{Layer 2: $\dh$, $\Delta \hat h$, $\dh^*$ }
    \label{fig:layer2_dh}
    \end{subfigure}
    \begin{subfigure}[b]{0.235\textwidth}
    \centering
    \includegraphics[width=\textwidth]{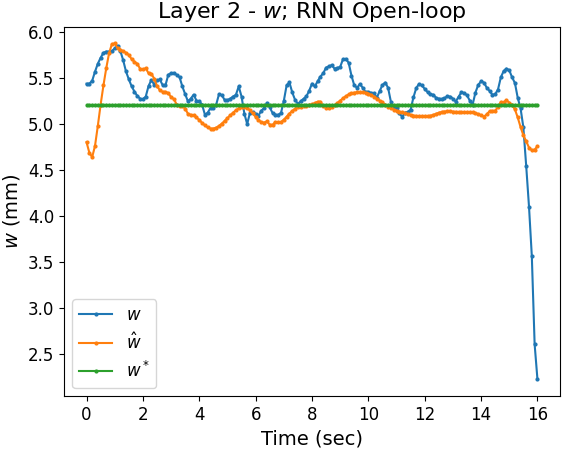}
    \caption{Layer 2: $w$, $\hat w$, $w^*$}
    \label{fig:layer2_w}
    \end{subfigure}
    \begin{subfigure}[b]{0.235\textwidth}
    \centering
    \includegraphics[width=\textwidth]{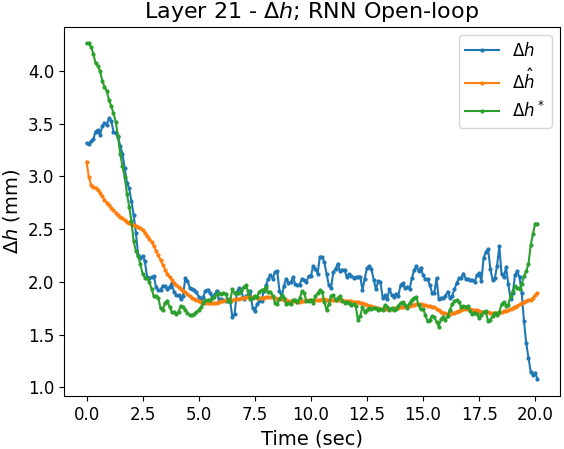}
    \caption{Layer 21: $\dh$, $\Delta \hat h$, $\dh^*$ }
    \label{fig:layer21_dh}
    \end{subfigure}
    \begin{subfigure}[b]{0.235\textwidth}
    \centering
    \includegraphics[width=\textwidth]{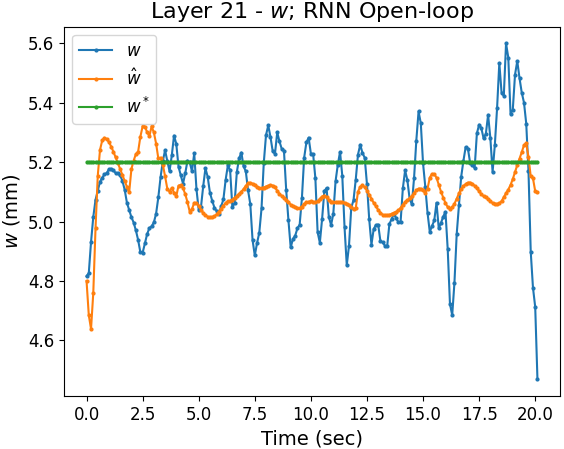}
    \caption{Layer 21: $w$, $\hat w$, $w^*$}
    \label{fig:layer21_w}
    \end{subfigure}
    \caption{Actual output, estimated output, and target output for Layer 2 and Layer 21 (next to the last layer)}
    \label{fig:layer2_21}
\end{figure}

\begin{figure}[h!]
    \centering
    \begin{subfigure}[b]{0.235\textwidth}
    \centering
    \includegraphics[width=\textwidth]{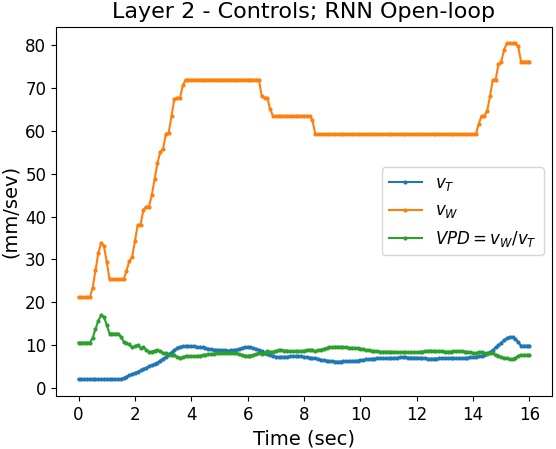}
    \caption{Layer 2: $v_T$, $v_W$, VPD}
    \label{fig:layer2_u}
    \end{subfigure}
    \begin{subfigure}[b]{0.235\textwidth}
    \centering
    \includegraphics[width=\textwidth]{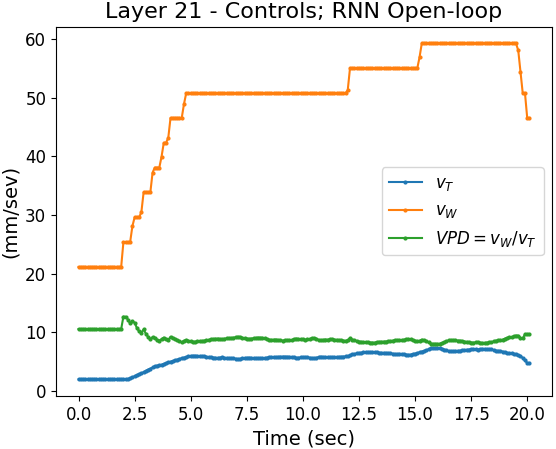}
    \caption{Layer 21: $v_T$, $v_W$, VPD}
    \label{fig:layer21_u}
    \end{subfigure}
    \caption{Input $(v_T,v_W)$ and VPD ($v_W/v_T$)}
    \label{fig:layer2_21_u}
\end{figure}

Figure~\ref{fig:layer2_21}  shows the deposition height and width of the layer 2 and layer 21, which is the next-to-the-final layer using the RNN model controller so we have a comparison between a lower layer and a higher layer. The sharp drop off at the end of the layer means a larger initial $\dh^*$ in the next layer, as shown in Figure~\ref{fig:layer2_dh} and \ref{fig:layer21_dh}.  Large $\dh^*$ means lower $v_T$. To avoid excessive VPD that will cause $w$ deviation, $v_W$ needs to also drop.  This is seen in both the initial portion of Figure~\ref{fig:layer2_u} and \ref{fig:layer21_u}.  However, in this case, $v_W$ reaches its lower bound, resulting in a jump in VPD.  This may results in larger $w$ as seen in Figure~\ref{fig:layer2_w}.  After the initial compensation for the drop off in the previous layer, the predicted output $(\Delta \hat h, \hat w)$ follow the desired output $(\Delta h^*,w^*)$ closely, demonstrating that the one-step-ahead controller is working reasonably well.  However, the actual output $(\dh,w)$ show larger error, especially near the end of the layer, showing the predictive model should be further improved. Furthermore, comparing the model predictions between layer 2 and layer 21, the model predicts the higher layer deposition better than the lower layer. As we mentioned earlier, the cooling rate at the higher layer is expected to be lower than that at the lower layer as the layers build up away from the heat sink. The prediction error deviation may come from different thermal conditions and the model being trained leaning toward higher temperatures.
There is also a sharp increase of $\dh^*$ at the end of the layer, but the physical system is unable to produce target deposition and the sharp drop off remains.  Note that the controller strives to keep VPD constant which is essential to regulate both $\dh$ and $w$.  When $v_W$ saturates at the lower bound, VPD jumps and the performance degrades.


\begin{figure}[h!]
\centering
    \begin{subfigure}[b]{0.23\textwidth}
    \centering
    \includegraphics[width=1\textwidth]{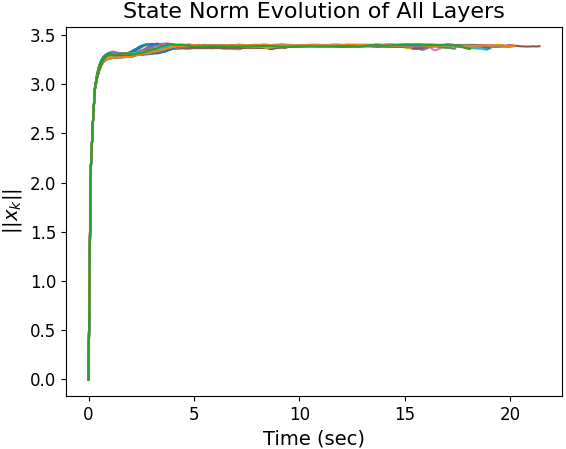}
    \caption{State norm evolution}
    \label{fig:layers_state_norm_evo}
    \end{subfigure}
    \begin{subfigure}[b]{0.24\textwidth}
    \centering
    \includegraphics[width=\textwidth]{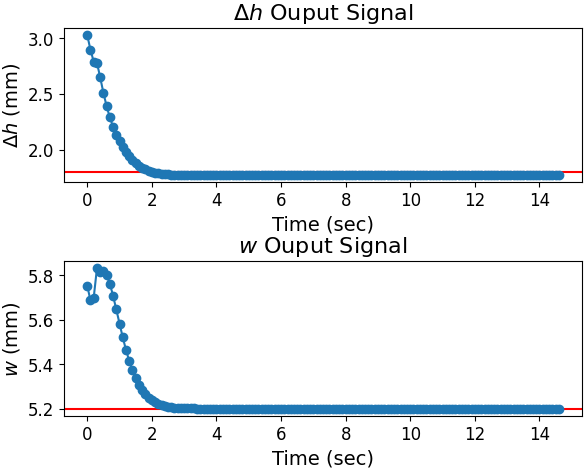}
    \caption{Output convergence}
    \label{fig:constantu}
    \end{subfigure}
    \caption{Norm of state evolution for all layers and output convergence under constant $u^*$ }
    \label{fig:transientresponse}
\end{figure}

\begin{figure}[h!]
\centering
\includegraphics[width=0.5\linewidth]{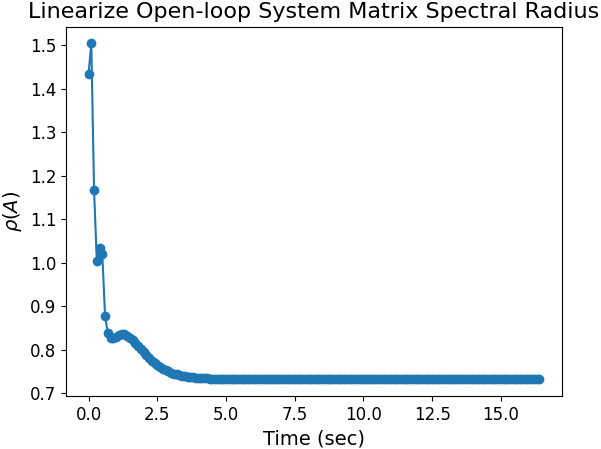}
    \caption{Spectral Radius of the linearized system along the state trajectory}
    \label{fig:A_matrix}
\end{figure}

The initial state $x_0$ of each layer is set to the zero vector. This produces a transient as shown in Figure~\ref{fig:layers_state_norm_evo} which includes the evolution of the norm of the state vector in each layer.  The state vector jumps from zero quickly, in less than a second.  The overall state transient decays in about 4~seconds which is about the same initial transient period we observed at the starting edge. As an additional stability check, we feed the constant input in the baseline case, $u^*$, into the simple RNN model.  In Figure~\ref{fig:constantu}, the output converges close to the chosen target output $y^*$, showing the model stability in the steady state.  
For the linearized open loop system, the spectral radius of the system matrix is initially larger than 1 (see Figure~\ref{fig:A_matrix}), showing potential local instability.  This corresponds to the period where the deposit builds up quickly.  After this initial transient for about 2~seconds, the linearized system becomes stable.

\subsection{Thermal Analysis}

\begin{figure}[htbp]
     \centering
     \begin{subfigure}[b]{0.235\textwidth}
         \centering
         \includegraphics[width=\textwidth]{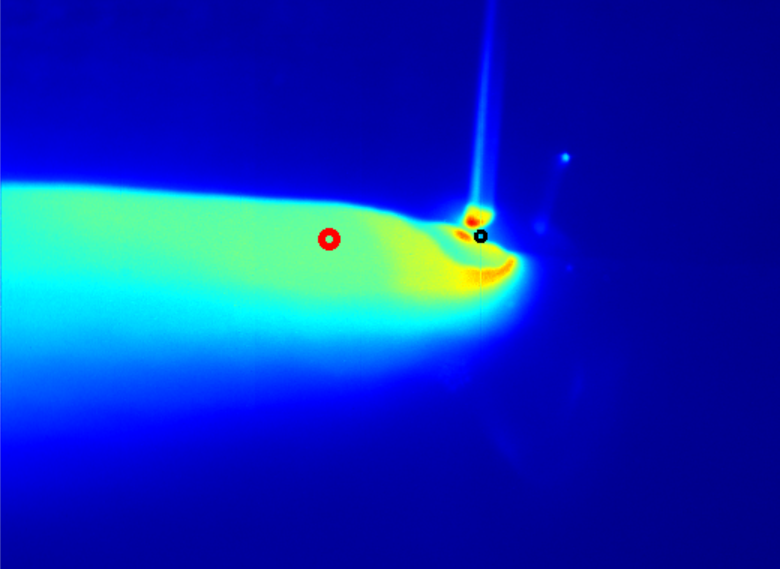}
         \caption{Thermal field visualization.}
         \label{fig:bead_thermal_sample}
     \end{subfigure}
     \begin{subfigure}[b]{0.235\textwidth}
         \centering
         \includegraphics[width=\textwidth]{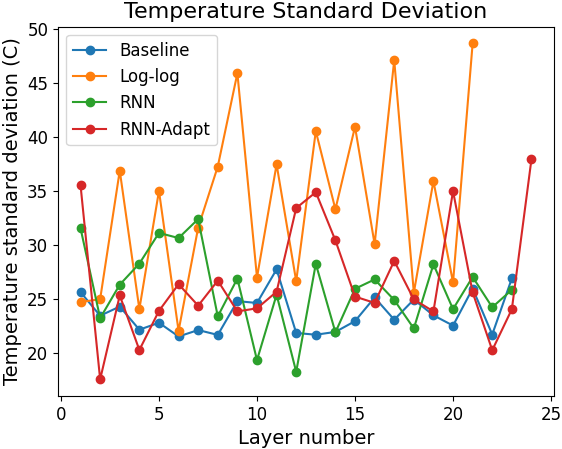}
         \caption{Thermal profile SD.}
         \label{fig:thermal_std}
     \end{subfigure}
    \caption{Thermal profile standard deviation across layers for the four controller implementations. Temperature is evaluated within the region indicated by the red circle.}
    \label{fig:thermal_std_layers}
\end{figure}

Previous studies have reported a positive correlation between bead width and infrared temperature intensity in WAAM processes~\cite{he2025waaminfrared}. To further examine the physical consistency, we analyze the thermal behavior under each controller. Temperature measurements are obtained using a Xiris XIR-1800 thermal camera~\cite{Xiris}, mounted on another robot to follow the torch during deposition. As shown in Fig.~\ref{fig:bead_thermal_sample}, a $1\,\text{mm} \times 1\,\text{mm}$ region located 8~mm behind the torch is selected. The average temperature within this region is computed at each time step, and the standard deviation of this temperature signal is calculated across each layer to quantify thermal consistency.

Figure~\ref{fig:thermal_std} presents the standard deviation of the temperature across layers for all controller implementations, and Table~\ref{tab:thermal_std} summarizes the layer-averaged results. The log–log controller, which exhibits larger bead width variation, also shows significantly higher thermal fluctuation. In contrast, the baseline and RNN-based controller demonstrate lower and more stable thermal variation. Comparing the RNN and RNN-Adaptive controllers, the RNN-Adaptive controller exhibits slightly higher width variation and thermal variation. These observations suggest a strong correlation between geometric consistency and thermal stability. In particular, reduced width variation is accompanied by reduced thermal fluctuation. This further indicates that thermal dynamics play an important role in WAAM geometric quality.

\begin{table}[ht]
\begin{center}
\begin{tabular}{ c c c c } 
\hline
\hline
 Baseline & Log-log & RNN & RNN-Adaptive  \\
\hline
 23.6 & 33.4 & 25.9 & 26.7 \\
\hline
\hline
\end{tabular}
\caption{Average thermal profile standard deviation across all layers. Unit: °C}
\label{tab:thermal_std}
\end{center}
\end{table}

\subsection{Adaptive Fine-tuning Model Prediction Performance}

\begin{figure}[htbp]
     \centering
     \begin{subfigure}[b]{0.235\textwidth}
         \centering
         \includegraphics[width=\textwidth]{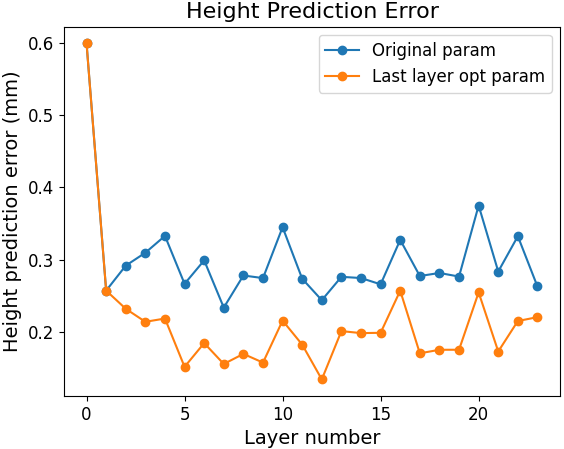}
         \caption{Height prediction error.}
         \label{fig:height_error_ada_full_online}
     \end{subfigure}
     \begin{subfigure}[b]{0.235\textwidth}
         \centering
         \includegraphics[width=\textwidth]{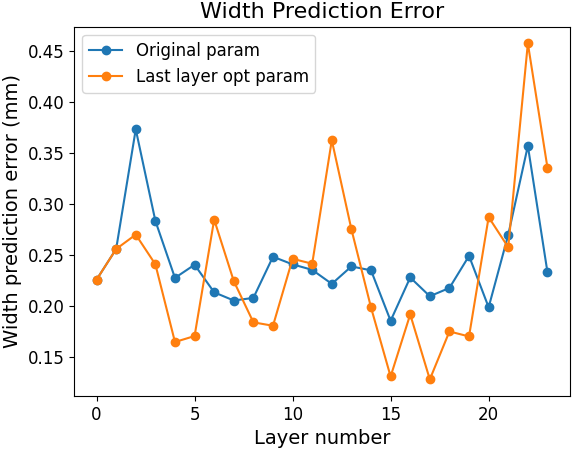}
         \caption{Width prediction error}
         \label{fig:width_error_ada_full_online}
     \end{subfigure}
    \caption{Comparison of the average height and width prediction errors between the original trained model and the model with adaptive fine-tuning.}
    \label{fig:hw_error_adaptive}
\end{figure}

\begin{figure}[h!]
\centering
    \begin{subfigure}[b]{0.235\textwidth}
    \centering
    \includegraphics[width=\textwidth]{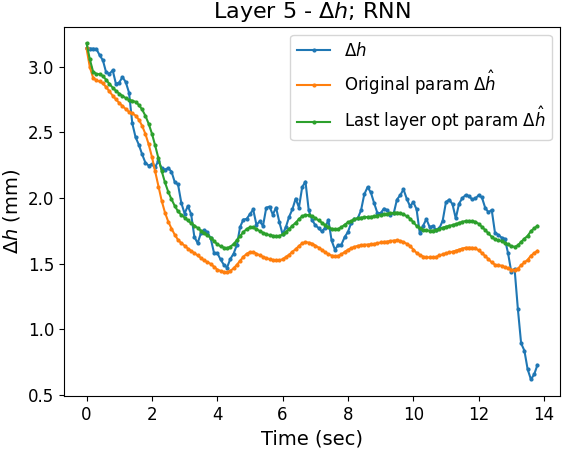}
    \caption{Layer 5: $\dh$ }
    \label{fig:layer_5_dh_adapt}
    \end{subfigure}
    \begin{subfigure}[b]{0.235\textwidth}
    \centering
    \includegraphics[width=\textwidth]{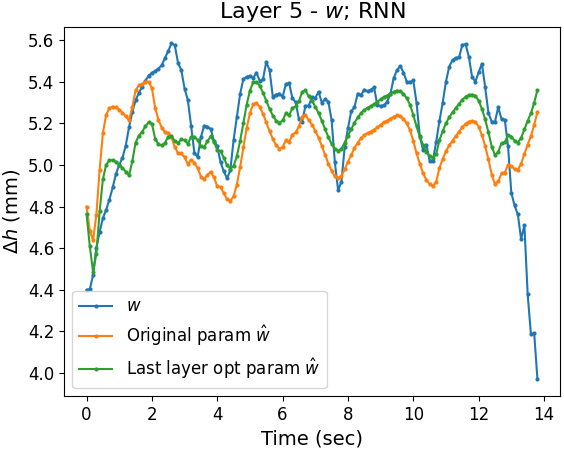}
    \caption{Layer 5: $w$}
    \label{fig:layer_5_w_adapt}
    \end{subfigure}
    \begin{subfigure}[b]{0.235\textwidth}
    \centering
    \includegraphics[width=\textwidth]{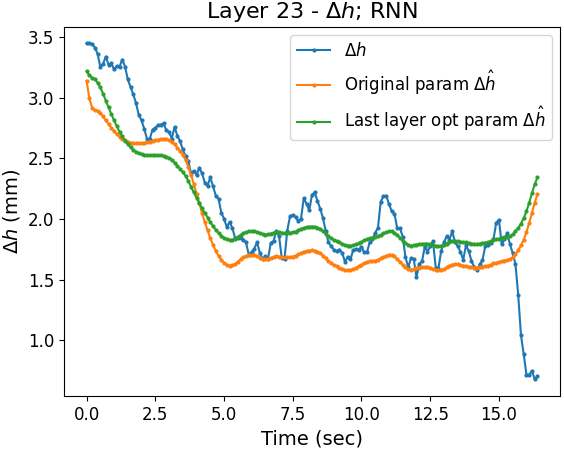}
    \caption{Layer 23: $\dh$}
    \label{fig:layer_23_dh_adapt}
    \end{subfigure}
    \begin{subfigure}[b]{0.235\textwidth}
    \centering
    \includegraphics[width=\textwidth]{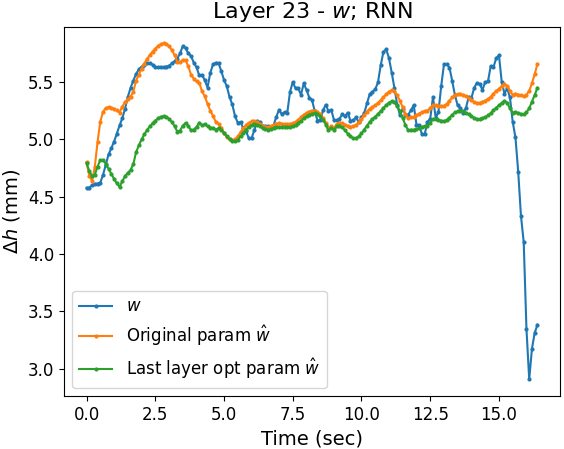}
    \caption{Layer 23: $w$}
    \label{fig:layer_23_w_adapt}
    \end{subfigure}
    \caption{Comparison among the measured deposition profile, the prediction from the original trained model, and the prediction from the adaptively fine-tuned model. The adaptively fine-tuned model improves the height prediction, while the width prediction improves for some layers but remains less consistent.}
    \label{fig:layer_hw_adaptive}
\end{figure}

Figure~\ref{fig:layer2_21} compares the prediction from the original trained model with the measured deposition. Although the original model predicts the deposition reasonably well, a noticeable discrepancy remains between the predicted and measured profiles. To evaluate whether adaptive fine-tuning improves model accuracy, the data collected under the RNN-Adaptive controller setting are used to compare predictions from the original trained model and the adaptively fine-tuned model at each layer. Figure~\ref{fig:hw_error_adaptive} compares the average height and width prediction errors between the original model and the adaptive fine-tuning model using previous-layer data. The results show that adaptive fine-tuning significantly improves the height prediction accuracy. For width prediction, adaptive fine-tuning improves the prediction in some layers, while in other layers its performance is comparable to or slightly worse than that of the original model.  This suggests that the model structure for the width dynamics will need to be improved in the future.
Figure~\ref{fig:layer_hw_adaptive} shows representative layer-wise comparisons among the measured deposition, the prediction from the original model, and the prediction from the adaptively fine-tuned model. Overall, adaptive fine-tuning enables the model to better adapt to local process conditions that are not fully captured by the original offline training data, especially for height prediction.

\section{Conclusion}
\label{conclusion}

This paper formulates the WAAM process as an input–output dynamical system, with torch speed and wire feed rate as inputs and bead height and width as outputs. Data-driven approaches based on RNNs are employed to learn the process dynamics. 
An one-step-ahead-ahead predictive control strategy is then developed to regulate deposition geometry using the learned models. 
Closed-loop control experiments demonstrate that the proposed RNN-based control framework improves welding geometric consistency compared with static-input baselines and analytical models. We  analyze the temperature profile of each controller. The results show with higher width variation, there is a higher temperature variation as well. 

Future work will investigate incorporating additional process factors into the modeling and control framework, particularly thermal information such as weld bead temperature, to further enhance prediction and control performance. In addition, conventional welding often employs weaving motions to improve multi-pass and multi-layer deposition quality. Extending the current framework to include additional motion degrees of freedom, such as controlled torch translation and orientation modulation, may further improve geometric consistency and overall weld quality.

\appendix

\subsection{RNN and NN-NARX Architectures}
\label{app:RNN}
\subsubsection{Simple RNN}

The RNN model is described by
\begin{equation}
    \begin{aligned}
        \x{k+1} &= \tanh(W_{ih}\u k + W_{hh}\x k + b_h), \\
        \y k &= W_{ho}\x k + b_o,
    \end{aligned}
\end{equation}
where $\x k$ denotes the hidden state, $\u k$ is the input to the model, $W$ and $b$ represent the weight matrices and bias vectors, respectively, and $\tanh(\cdot)$ is the hyperbolic tangent activation function. The recurrent weight matrix $W_{hh}$ captures temporal dependencies in the hidden state, while $W_{ih}$ and $W_{ho}$ regulate the influence of the input and the projection to the output.

\subsubsection{LSTM}

The LSTM model introduces gating mechanisms to better capture long-term dependencies and mitigate gradient vanishing issues. Its dynamics are given by
\begin{equation}
    \begin{aligned}
        i_{k+1} &= \sigma(W_{ii}\u k + W_{hi}\x k + b_i), \\
        f_{k+1} &= \sigma(W_{if}\u k + W_{hf}\x k + b_f), \\
        g_{k+1} &= \tanh(W_{ig}\u k + W_{hg}\x k + b_g), \\
        o_{k+1} &= \sigma(W_{io}\u k + W_{ho}\x k + b_o), \\
        c_{k+1} &= f_{k+1} \odot c_{k} + i_{k+1} \odot g_{k+1}, \\
        \x{k+1} &= o_{k+1} \odot \tanh(c_{k+1}), \\
        \y k &= W_{yh}\x k + b_y,
    \end{aligned}
\end{equation}
where $\sigma(\cdot)$ denotes the sigmoid activation function and $\odot$ denotes the Hadamard product. The input $i_t$, forget $f_t$, and output $o_t$ gates regulate information flow into and out of the cell state $c_t$, allowing the model to represent longer-term temporal dependencies.

\subsubsection{GRU}

The GRU simplifies the LSTM architecture while retaining gating mechanisms. Its dynamics are defined as
\begin{equation}
    \begin{aligned}
        z_{k+1} &= \sigma(W_{iz}\u k + W_{hz}\x k + b_z), \\
        r_{k+1} &= \sigma(W_{ir}\u k + W_{hr}\x k + b_r), \\
        \tilde{h}_{k+1} &= \tanh(W_{ih}\u k + W_{hh}(r_k \odot \x k) + b_h), \\
        \x{k+1} &= (1 - z_{k+1}) \odot \x k + z_{k+1} \odot \tilde{h}_{k+1}, \\
        \y k &= W_{yh}\x k + b_y,
    \end{aligned}
\end{equation}
where the reset gate $r_t$ controls the influence of past hidden states and the update gate $z_t$ determines how much new information is incorporated into the hidden state.

\subsubsection{NN-NARX}

In contrast to recurrent models, the NN-NARX model does not maintain a hidden state that propagates through time. Instead, it directly maps a stack of past inputs and outputs to the current output:
\begin{equation}
    \begin{aligned}
        x_1 &= \phi(W_1 {\underline u}_k + b_1), \\
        x_2 &= \phi(W_2 x_1 + b_2), \\
        \y k &= W_3 x_2 + b_3,
    \end{aligned}
\end{equation}
where ${\underline u}_k  = [y_{k-1}, y_{k-2}, \dots, u_k, u_{k-1}, \dots]^T$ is the regressor vector and $\phi(\cdot)$ denotes a nonlinear activation function. The length of the regression vector is determined by the choice of the system order. 



\bibliographystyle{IEEEtran}
\bibliography{bib}

@INPROCEEDINGS{Dharmawan2020model,
  author={Dharmawan, Audelia G. and Xiong, Yi and Foong, Shaohui and Song Soh, Gim},
  booktitle={2020 IEEE International Conference on Robotics and Automation (ICRA)}, 
  title={A Model-Based Reinforcement Learning and Correction Framework for Process Control of Robotic Wire Arc Additive Manufacturing}, 
  year={2020},
  venue 	= {Paris, France},
  eventdate = {May 31 –- June 4},
  volume={},
  number={},
  pages={4030-4036},
  doi={10.1109/ICRA40945.2020.9197222}}

@article{Xiong2019process,
author = {Xiong, Yi and others},
year = {2019},
month = {12},
pages = {4159-4170},
title = {Process planning for adaptive contour parallel toolpath in additive manufacturing with variable bead width},
volume = {105},
journal = {International Journal of Advanced Manufacturing Technology},
doi = {10.1007/s00170-019-03954-1}
}

@article{Handa1997THERE,
  title={The robotic easy teaching system in computer aided welding},
  author={Handa, H and Okumura, S and Nio, S},
  journal={NIST Special Publication(USA),},
  volume={923},
  pages={562--575},
  year={1997}
}

@article{he2025open,
title = {Open-source software architecture for multi-robot Wire Arc Additive Manufacturing (WAAM)},
journal = {Applications in Engineering Science},
volume = {22},
pages = {100204},
year = {2025},
issn = {2666-4968},
doi = {https://doi.org/10.1016/j.apples.2025.100204},
author = {Honglu He and others}
}

@article{lu2025multi,
    author = {Lu, Chen-Lung and others},
    title = {Multi-Robot Scan-n-Print for Wire Arc Additive Manufacturing},
    journal = {ASME Letters in Translational Robotics},
    volume = {1},
    number = {1},
    pages = {011003},
    year = {2025},
    month = {02},
    doi = {10.1115/1.4067825}
}

@article{Kingma2014AdamAM,
  title={Adam: A Method for Stochastic Optimization},
  author={Diederik P. Kingma and Jimmy Ba},
  journal={CoRR},
  year={2014},
  volume={abs/1412.6980}
}

@Article{wacker2021geometry,
AUTHOR = {Wacker, Christian and others},
TITLE = {Geometry and Distortion Prediction of Multiple Layers for Wire Arc Additive Manufacturing with Artificial Neural Networks},
JOURNAL = {Applied Sciences},
VOLUME = {11},
YEAR = {2021},
NUMBER = {10},
ARTICLE-NUMBER = {4694},
ISSN = {2076-3417},
DOI = {10.3390/app11104694}
}

@article{kaikui2023prediction,
  author    = {Kaikui Zheng and Chuanxu Yao and Gang Mou and Hongliang Xiang},
  title     = {Prediction of Weld Bead Formation of Duplex Stainless Steel Fabricated by Wire Arc Additive Manufacturing Based on the PSO-BP Neural Network},
  journal   = {Journal of Marine Science and Application},
  volume    = {22},
  number    = {2},
  pages     = {311--323},
  year      = {2023},
  month     = jun,
  doi       = {10.1007/s11804-023-00332-y},
  issn      = {1993-5048}
}

@Article{so2024prediction,
AUTHOR = {So, Min Seop and Mahdi, Mohammad Mahruf and Kim, Duck Bong and Shin, Jong-Ho},
TITLE = {Prediction of Metal Additively Manufactured Bead Geometry Using Deep Neural Network},
JOURNAL = {Sensors},
VOLUME = {24},
YEAR = {2024},
NUMBER = {19},
ARTICLE-NUMBER = {6250},
PubMedID = {39409290},
ISSN = {1424-8220},
DOI = {10.3390/s24196250}
}

@Manual{Xiris,
title = {XIR-1800},
author = {{Xiris}},
year = {2026},
url = {https://info.xiris.com/xir-1800-thermal-camera/},
}

@article{WU2018review,
title = {A review of the wire arc additive manufacturing of metals: properties, defects and quality improvement},
journal = {Journal of Manufacturing Processes},
volume = {35},
pages = {127-139},
year = {2018},
issn = {1526-6125},
doi = {https://doi.org/10.1016/j.jmapro.2018.08.001},
author = {Bintao Wu and others}
}

@article{xiong2014bead,
  title={Bead geometry prediction for robotic GMAW-based rapid manufacturing through a neural network and a second-order regression analysis},
  author={Xiong, Jun and Zhang, Guangjun and Hu, Jianwen and Wu, Lin},
  journal={Journal of Intelligent Manufacturing},
  volume={25},
  number={1},
  pages={157--163},
  year={2014},
  publisher={Springer}
}

@article{mu2022layer,
  title={Layer-by-layer model-based adaptive control for wire arc additive manufacturing of thin-wall structures},
  author={Mu, Haochen and Polden, Joseph and Li, Yuxing and He, Fengyang and Xia, Chunyang and Pan, Zengxi},
  journal={Journal of Intelligent Manufacturing},
  volume={33},
  number={4},
  pages={1165--1180},
  year={2022},
  publisher={Springer}
}

@article{xiong2016closed,
  title={Closed-loop control of variable layer width for thin-walled parts in wire and arc additive manufacturing},
  author={Xiong, Jun and Yin, Ziqiu and Zhang, Weihua},
  journal={Journal of Materials Processing Technology},
  volume={233},
  pages={100--106},
  year={2016},
  publisher={Elsevier}
}

@article{xia2020model,
  title={Model predictive control of layer width in wire arc additive manufacturing},
  author={Xia, Chunyang and others},
  journal={Journal of Manufacturing Processes},
  volume={58},
  pages={179--186},
  year={2020},
  publisher={Elsevier}
}

@inproceedings{mendes2025data,
  title={Data-Driven Modeling and Closed-Loop Control of Bead Geometry in Wire Arc Additive Manufacturing},
  author={Mendes, Marcel and Novais, Igor and Lizarralde, Fernando},
  booktitle={2025 International Joint Conference on Neural Networks (IJCNN)},
  pages={1--8},
  year={2025},
  organization={IEEE}
}

@article{wang2022control,
  title={Control of bead geometry using multiple model approach in wire-arc additive manufacturing (WAAM)},
  author={Wang, Zeya and Zimmer-Chevret, Sandra and L{\'e}onard, Fran{\c{c}}ois and Abba, Gabriel},
  journal={The International Journal of Advanced Manufacturing Technology},
  volume={122},
  number={7},
  pages={2939--2951},
  year={2022},
  publisher={Springer}
}

@article{mattera2025optimal,
  title={Optimal data-driven control of manufacturing processes using reinforcement learning: an application to wire arc additive manufacturing},
  author={Mattera, Giulio and Caggiano, Alessandra and Nele, Luigi},
  journal={Journal of Intelligent Manufacturing},
  volume={36},
  number={2},
  pages={1291--1310},
  year={2025},
  publisher={Springer}
}

@incollection{pyTorch,
  title = {PyTorch: An Imperative Style, High-Performance Deep Learning Library},
  author = {Paszke, Adam and others},
  booktitle = {Advances in Neural Information Processing Systems 32},
  editor = {H. Wallach and H. Larochelle and A. Beygelzimer and d'Alch\'{e}-Buc, F. and Fox, E. and Garnett, R.},
  pages = {8024--8035},
  year = {2019},
  publisher = {Curran Associates, Inc.}
}

@article{he2025waaminfrared,
    author = {He, Honglu and Wen, John T.},
    title = {Wire Arc Additive Manufacturing With Infrared Image Feedback},
    journal = {ASME Letters in Translational Robotics},
    volume = {1},
    number = {1},
    pages = {011004},
    year = {2025},
    month = {02},
    issn = {2997-9765},
    doi = {10.1115/1.4067980}
}

@article{simpleRNN,
  title={Finding structure in time},
  author={Elman, Jeffrey L},
  journal={Cognitive science},
  volume={14},
  number={2},
  pages={179--211},
  year={1990},
  publisher={Wiley Online Library}
}

@article{lstm,
  title={Long short-term memory},
  author={Hochreiter, Sepp and Schmidhuber, J{\"u}rgen},
  journal={Neural computation},
  volume={9},
  number={8},
  pages={1735--1780},
  year={1997},
  publisher={MIT Press}
}

@article{gru,
  title={Empirical evaluation of gated recurrent neural networks on sequence modeling},
  author={Chung, Junyoung and Gulcehre, Caglar and Cho, KyungHyun and Bengio, Yoshua},
  journal={arXiv preprint arXiv:1412.3555},
  year={2014}
}

\end{document}